\documentclass[article, 10 pt, conference]{ieeeconf}  
\IEEEoverridecommandlockouts                              

\UseRawInputEncoding                                    

\let\labelindent\relax
\usepackage{enumitem}
\setlist{topsep=1pt, partopsep=0pt, parsep=0pt, itemsep=0pt}
\usepackage[
top    = 0.75in,
bottom = 0.75in,
left   = 0.75in,
right  = 0.75in]{geometry}
\usepackage{cite}
\usepackage[skip=10pt plus1pt, indent=35pt]{parskip}
\usepackage{placeins}
\usepackage{textcomp}
\usepackage[dvipsnames]{xcolor}
\definecolor{royalblue}{RGB}{65, 105, 225}
\definecolor{maroon}{RGB}{180, 0, 0}
\definecolor{DarkGreen}{RGB}{0, 100, 0}
\usepackage{booktabs}
\usepackage{multirow}
\usepackage{siunitx}
\usepackage{graphicx}
\usepackage[symbol]{footmisc}
\usepackage{adjustbox}
\usepackage{float}
\usepackage{svg}
\usepackage{array}
\usepackage[T1]{fontenc}
\usepackage{subcaption}
\usepackage{xspace}
\usepackage{amsmath}
\usepackage{scrextend}
\usepackage{caption}
\usepackage{tabularx}
\usepackage{listings}
\usepackage{wrapfig}
\usepackage{calligra}
\usepackage{comment}
\usepackage{soul}
\usepackage{balance}
\usepackage{booktabs}
\usepackage{graphicx}
\usepackage{array}
\usepackage{listings}
\usepackage{xcolor}
\usepackage{caption}
\newcolumntype{A}{ >{\centering\arraybackslash} m{4cm} }
\newcolumntype{B}{ >{\centering\arraybackslash} m{1cm} }
\newcolumntype{C}[1]{>{\centering\let\newline\\\arraybackslash\hspace{0pt}}m{#1}}

\makeatletter
\newcommand\footnoteref[1]{\protected@xdef\@thefnmark{\ref{#1}}\@footnotemark}
\makeatother

\usepackage{amsmath,lipsum} 
\usepackage{amssymb}  
\usepackage{amsfonts}
\usepackage{textgreek}
\usepackage{authblk}
\usepackage{etoolbox}

\makeatletter
\let\NAT@parse\undefined

\let\oldthebibliography\thebibliography
\let\endoldthebibliography\endthebibliography
\renewenvironment{thebibliography}[1]{
  \oldthebibliography{#1}
  \setlength{\itemsep}{-1.75ex plus-.1ex minus-.1ex} 
}{
  \endoldthebibliography
}

\makeatother

\usepackage[colorlinks=true, citecolor=cyan]{hyperref}  
\def\BibTeX{{\rm B\kern-.05em{\sc i\kern-.025em b}\kern-.08em
    T\kern-.1667em\lower.7ex\hbox{E}\kern-.125emX}}

\title{\bf
Anticipate, Adapt, Act: A Hybrid Framework for Task Planning
}
\author{Nabanita Dash$^1$, Ayush Kaura$^1$, Shivam Singh$^1$, Ramandeep Singh$^1$, Snehasis Banerjee$^2$, \\ Mohan Sridharan$^3$,  K. Madhava Krishna$^1$

\thanks{$^{1}$ Robotics Research Center, IIIT Hyderabad, India}
\thanks{$^{2}$ TCS Research, Tata Consultancy Services, India}
\thanks{$^{3}$ School of Informatics, University of Edinburgh, UK}
}

\makeatletter
\renewcommand{\@seccntformat}[1]{%
  \protect\csname the#1\endcsname\protect\quad%
}
\makeatother

\renewcommand{\thesection}{\arabic{section}}
\renewcommand{\thesubsection}{\thesection.\arabic{subsection}}
\renewcommand{\thesubsubsection}{\thesubsection.\arabic{subsubsection}}

\begin{document}

\maketitle
\thispagestyle{empty}
\pagestyle{empty}

\begin{abstract}

Anticipating and adapting to failures is a key capability robots need to collaborate effectively with humans in complex domains. This continues to be a challenge despite the impressive performance of state of the art AI planning systems and Large Language Models (LLMs) because of the uncertainty associated with the tasks and their outcomes. Toward addressing this challenge, we present a hybrid framework that integrates the generic prediction capabilities of an LLM with the probabilistic sequential decision-making capability of Relational Dynamic Influence Diagram Language. For any given task, the robot reasons about the task and the capabilities of the human attempting to complete it; predicts potential failures due to lack of ability (in the human) or lack of relevant domain objects; and executes actions to prevent such failures or recover from them. Experimental evaluation in the VirtualHome 3D simulation environment demonstrates substantial improvement in performance compared with state of the art baselines. 


\end{abstract}
\vspace{-1em}
\vspace{-0.7em}
\section{Introduction}  
\begin{figure}[t]
    \centering
    \includegraphics[width=1\linewidth]{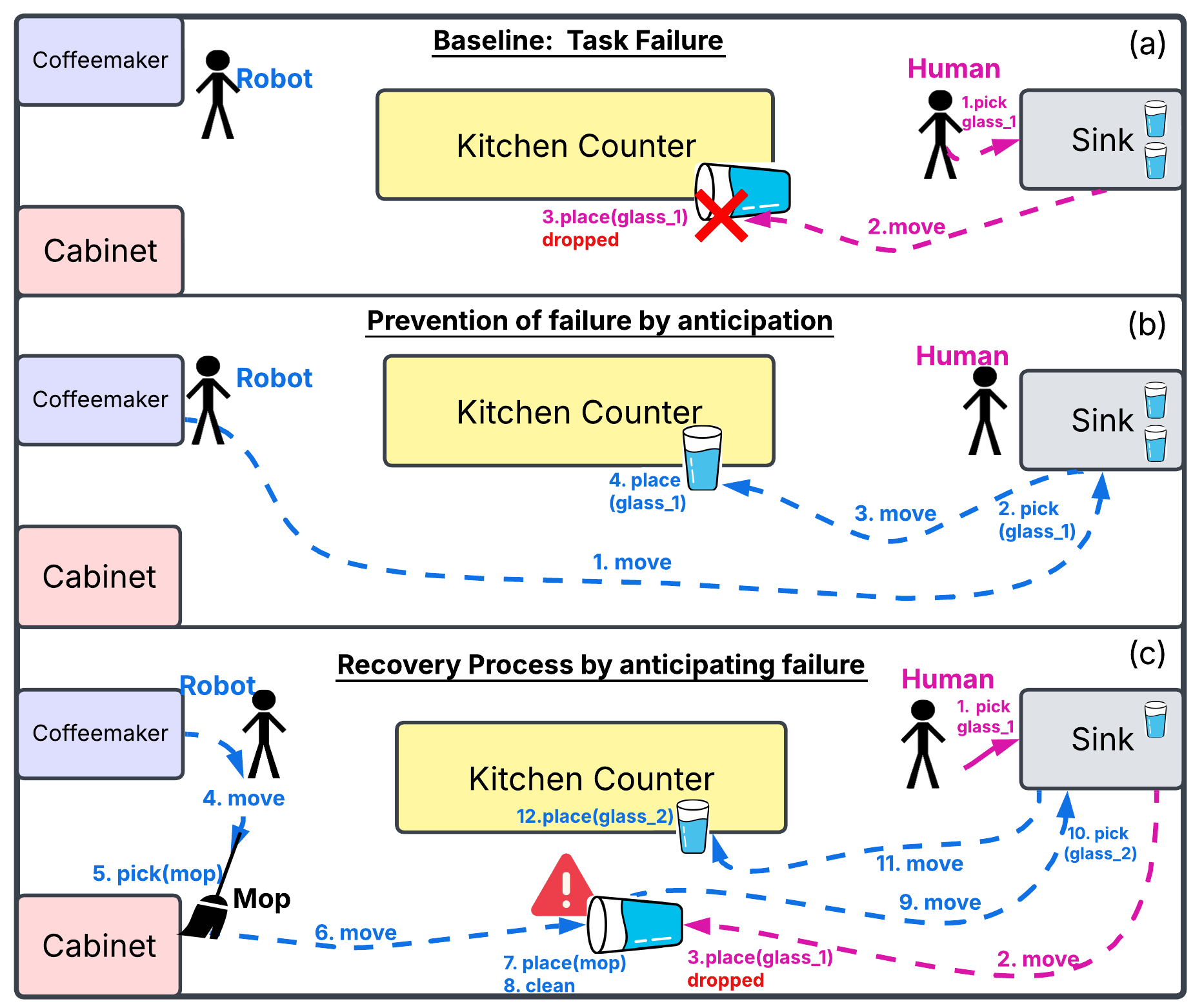}
    \caption{Illustrative task of fetching a glass of water from the sink to the kitchen counter. In the baseline scenario, the human may end up dropping the water glass due to stability issues. Our framework enables the robot to anticipate such failures; it either prevents the failure by completing the task, or prepares to recover from the failure by fetching a mop that can be used to clean the potential water spill.}
    \label{fig:teaser}
\vspace{-0.5cm}
\end{figure}

Consider a robot assisting an elderly human in a kitchen, say with fetching a glass of water from the sink to the kitchen counter. Due to mobility and stability limitations, there is uncertainty about whether the human can complete the task successfully; they may end up dropping the water glass. We expect the robot to anticipate the potential for such negative outcomes, i.e., the glass being dropped, and either prevent this negative outcome, e.g., by fetching the water glass, or prepare to deal with the outcome, e.g., by making sure it has access to the mop needed to clear the water spill. Figure~\ref{fig:teaser} shows some snapshots of these scenarios. State of the art methods for robot planning and  human-robot collaboration assume deterministic environments (e.g., with classical planners~\cite{Brafman_Tolpin_Wertheim_2024, Jiang2019}), or pre-compute and use reactive policies (e.g., with probabilistic planners~\cite{9560871}), and do not fully support the desired proactive decision-making behavior. 

\vspace{-0.5em}
\noindent
The hybrid framework\footnote{\url{https://dnabanita7.github.io/ECMR2025}} presented in this paper is inspired by the observation that the desired adaptive behavior needs the ability to anticipate tasks, identify potential failures while executing actions to complete these tasks with a human, and to plan actions that prevent failures or help recover from them. Specifically, the framework enables the robot to:
\begin{itemize}
\vspace{-0.3cm}
    \item Adapt a pretrained Large Language Model (LLM) to anticipate future tasks and convert a description of uncertainty in task outcomes to model parameters of a stochastic planning problem.
    
    
    \item Use the Relational Dynamic Influence Diagram Language (RDDL) to encode and solve the stochastic planning problem, computing a sequence of actions to jointly achieve the current and upcoming tasks.
    
    
    \item Encode and reason with a reward mechanism that trades-off task completion probability with the effort needed to prevent or recover from potential failures. 
    
    
\vspace{-0.3cm}
\end{itemize}
Our key contribution is the smooth integration of the complementary strengths of the three components, i.e., LLM-based task anticipation, RDDL-based probabilistic planning, and proactive failure handling. We experimentally evaluate our framework in the context of human-robot collaboration in household tasks in the realistic VirtualHome simulation environment. We demonstrate an increase in task completion accuracy and a reduction in the number of failures, leading to improved human-robot collaboration compared with baselines that just use an LLM or a probabilistic planner.





\section{Related Work}
\vspace{-0.1cm}
There is an extensive body of research in Human-robot interaction (HRI), including recent advances in shared autonomy and collaborative task execution~\cite{Park2024, Chen2024}. Despite impressive advancements in perception, reasoning, and learning, adaptation to failures and collaboration between humans and robots continue to be open problems~\cite{tamp-survey,failure-resolution-model}. 

\vspace{-0.5em}
\noindent
Reasoning tasks such as planning and diagnostics have often been addressed by encoding prior domain knowledge as relational logic statements in an action language such as Planning Domain Definition Language (PDDL)~\cite{aeronautiques1998pddl} and using suitable solvers. Other languages such as RDDL~\cite{Sanner:RDDL} help model a class problems that are difficult to model with probabilistic extensions of PDDL (e.g, due to stochastic effects and unrestricted concurrency). Its semantics are that of a ground Dynamic Bayesian Network, and it can be used for both classical planning and probabilistic sequential decision making (e.g., Markov Decision Process, MDP; Partially Observable MDP, POMDP). It supports both classical tree search planners like PROST~\cite{keller-eyerich-icaps2012} and learning-based approaches in RDDL-Gym.

\vspace{-0.5em}
\noindent
In an attempt to reduce the effort involved in encoding domain knowledge, recent research has explored the use of data-driven frameworks such as LLMs for computing plans~\cite{10.5555/3666122.3667509}. The ability of LLMs to predict action sequences to complete tasks has led to claims about their ability to plan and reason~\cite{hirsch2024whatsplanevaluatingdeveloping, joublin2025copalcorrectiveplanningrobot}, although they do not build the models needed for reasoning and their operation does not match the directed search operation involved in planning. There is increasing experimental evidence demonstrating the tendency of LLMs to provide arbitrary responses in novel situations~\cite{Valmeekam2024LLMsSC}, advocating their use in planning frameworks for auxiliary tasks such as knowledge translation~\cite{xie2023translatingnaturallanguageplanning}, task anticipation~\cite{10611164,fu:roman25}, and goal allocation~\cite{izquierdo2024plancollabnl}. 

\vspace{-0.5em}
\noindent
Robust Human-Robot Collaboration (HRC) requires the ability to deal with action failures. Planning methods can monitor and adapt to deviations in action outcomes~\cite{nau:book04} using behavior models encoded in PDDL domains~\cite{izquierdo2022improved} or probabilistic sequential decision making~\cite{yang:JIRS21}. In the context of probabilistic sequential decision making, existing methods support adaptation to changes in  the domain~\cite{mohan:JAIR19}, learned models~\cite{Karia_2024}, and human behavior~\cite{10.1145/3171221.3171256}. 


\vspace{-0.5em}
\noindent
Despite the existing work, the desired proactive behavior that anticipates failures, and either prevents them or prepares to recover from them, continue to be a problem of interest. We seek to address this problem by leveraging the complementary strengths of knowledge-based and data-driven systems. Specifically, our hybrid framework combines the generic prediction capability of LLM, the probabilistic planning capability of RDDL, and a reward mechanism to trade off between task completion and failure recovery.




\begin{figure*}[t]
  \centering
  \includegraphics[width=\textwidth]{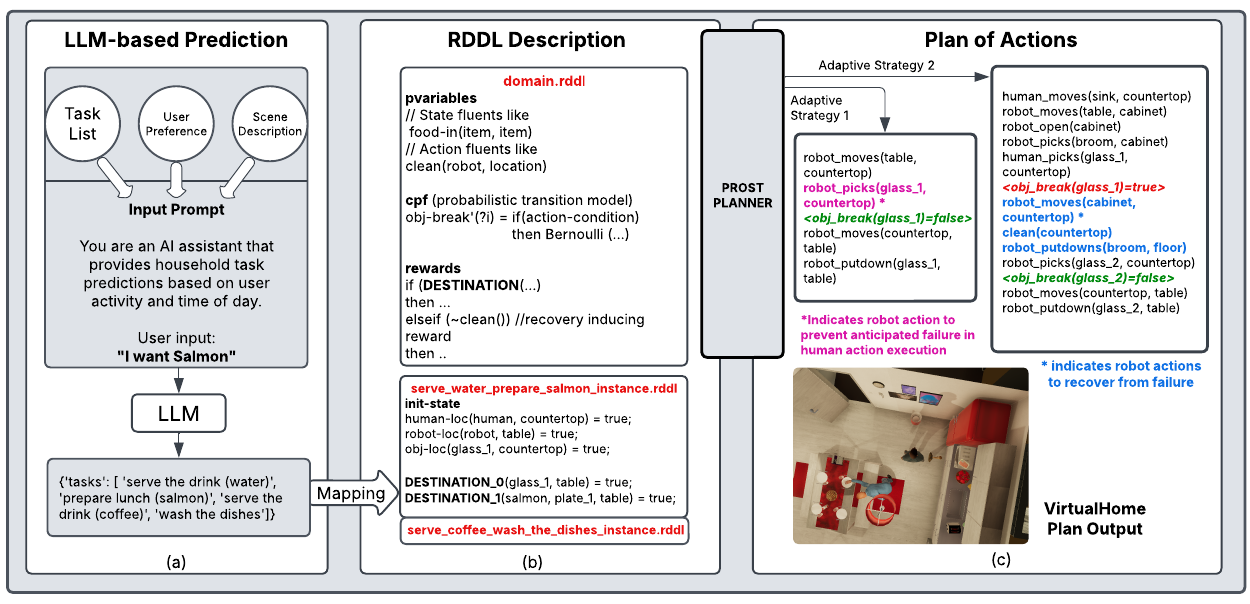}
  \vspace{-2em}
  \caption{Our framework's pipeline: (a) LLM takes a prompt of task lists, user preferences, scene description, and current command, to predict upcoming tasks; (b) RDDL description of domain knowledge and joint goals comprising current and predicted tasks fed to PROST planner; and (c) Plan of actions to be executed by robot and human to achieve the goal, including robot's actions to prevent or recover from potential failures due to the human's actions.}
  \label{fig:pipeline}
\vspace{-0.5cm}
\end{figure*}

\section{Problem Formulation and Framework}
\label{sec:formulation}
Consider a home environment with a human \( \mathcal{H} \) and an assistive robot \( \mathcal{R} \) collaborating to complete a given task specified as the goal \( G\). The sequence of high-level tasks \( \{T_1, T_2, \dots, T_n\}\) to be completed is not known to the robot in advance, and one task is normally assigned as \( G\) at a time. Completing each \( T_i \), e.g., \textit{preparing toast}, requires a plan of finer-granularity actions such as \textit{grab bread}, \textit{put-in toaster}, and \textit{switch appliance} to be computed and executed by the robot and the human. Completing a subset of these actions is considered a \textit{subgoal}.  
The execution of some actions can result in failure, e.g., a heavy plate with food on it may be dropped. Without loss of generality, we limit any such failure (in this paper) to the human's actions due to limitations in their capabilities, e.g., the human may not be able to lift heavy objects. For effective collaboration, the robot has to anticipate such failures based on prior knowledge or observations of the human's abilities. The robot can then prevent this failure, e.g., by completing the action instead of the human, or prepare to address this failure, e.g., fetch a broom and dustpan to clear the broken plate.  

\vspace{-0.5em}
\noindent
Figure~\ref{fig:pipeline} provides an overview of our hybrid framework for achieving the desired behavior. A robot equipped with this framework prompts an LLM with current task, prior knowledge of user preferences and the scene (if available), and some example task sequences, receiving as output a sequence of anticipated tasks (Section~\ref{sec:formulation-llm}). The current and next task are assigned as a joint goal to the domain-specific planning component (Section~\ref{sec:formulation-taskplan}). Assuming that the domain state is known after each action's execution (which is true in the simulation environment used for testing), domain-specific planning is formulated as a relational MDP, using RDDL to model domain-specific knowledge in the form of fluents, axioms, and a suitable reward structure. The output of this step is a plan of actions to be executed by the robot and the human, although the human's actual action choices may not match the robot's expectation. This planning also anticipates and accounts for failures (Section~\ref{sec:forumalation-anticipate}) to complete the tasks reliably. Specific components are described in detail below.

\vspace{-0.25em}
\subsection{LLM-based Task Anticipation}
\label{sec:formulation-llm}
In our framework, the \textit{llama-3.3-70b-versatile} LLM and the Groq API (temperature $0.2$, $\text{max\_tokens}=500$) is used to predict the next four tasks likely to be assigned. At run time, the prompt includes: (i) a \textit{task sample space} of 240 grounded household actions (\texttt{master\_tasks.json}); (ii) examples of user sequences
(\texttt{sequence.json}); and (iii) a scene graph of rooms and objects (\texttt{virtualhome\_categories.json}). The user sequences are sampled from the last three sequences completed by the user, and are updated across runs. During actual deployment, these sequences can be updated over time to adapt prediction to user's preferences.

\vspace{-0.5em}
\noindent
We used two prompting strategies: (i) few-shot and (ii) chain-of-thought~\cite{wei2023chainofthoughtpromptingelicitsreasoning}. Both strategies consider a predefined task space and a structured JSON description of the scene (often \textit{kitchen} in our experiments) including the locations of objects and other agents. Specifically, we considered 11 different tasks such as $move$ and $grab$ with multiple ground instantiations (e.g., move to different locations, grab different objects). The few-shot approach uses 2-3 prior observations of tasks completed by user, while the chain-of-thought method uses two in-context examples with step-by-step reasoning to infer user activity patterns. With either strategy, the LLM's output is a sequence of anticipated tasks, which is filtered to remove tasks considered to be invalid. 
A snapshot of such prompting and the corresponding output is shown in the left part of Figure~\ref{fig:pipeline}; the user asks for $salmon$ and the LLM predicts subsequent tasks to involve serving coffee and washing the dishes.

\subsection{Task Planning}
\label{sec:formulation-taskplan}
The JSON snippet is automatically mapped to the corresponding RDDL goal through simple template matching (\texttt{jsonfiles/rddl\_goals.json}). The current and the next (predicted) task (from LLM's output) are mapped to a joint goal \( G \). Recall that  domain-specific planning to achieve \(G\) is formulated as a relational MDP: \( \langle V, A, P, R, H, s_0 \rangle \), where \( V \) is the set of states, \( A \) is the set of finer-granularity actions (to be executed by robot or human), \( P \) is the state transition function, \( R \) is the reward specification, \( H \) is the planning horizon, and \( s_0 \) is the initial state. 

\vspace{-0.5em}
\noindent
In our RDDL domain description, each task \( T_i \) is automatically associated with an instance file defining relevant objects and axioms. These instance files are generated from a common domain file, encoding variables for states (e.g., location of objects, state of appliances) and actions,  universal transition dynamics, constraints, and reward structures that incorporate auxiliary incentives to guide the robot through intermediate steps for task completion. This approach supports modularity, with the domain file being defined once and the instance files defined based on the tasks at hand. In addition, subgoals are defined automatically as logically important stages on the path to any given goal, respecting necessary preconditions and dependencies between relvant states actions. Simulated trials validate these subgoals before they are encoded in the instance files, and used to evaluate partial achievement of the corresponding goal(s). 

\vspace{-0.5em}
\noindent
The reward function is designed to promote adaptive execution by trading off between successful task completion and the effort involved in preventing failures or recovering from them. Positive rewards are assigned for achieving subgoals and goals, while redundant or unsafe actions incur penalties. Additional details about reward specification are in Section~\ref{sec:forumalation-anticipate}, and an experimental analysis of reward sensitivity is in Figure~\ref{fig:rewards_failures_graph} in Section~\ref{subsec:exp_results}. The planner uses these rewards to generate a fine-grained action sequence by constructing a directed graph representation of possible states; it initializes Q-values to guide decision-making, checks reward locks, ensures that subgoal completion aligns with the overall goal, and prevents unnecessary delays.

\vspace{-0.5em}
\noindent
For computational efficiency, the original RDDL description is factored to obtain \( \langle D_R, D_H \rangle \), where \( D_R = \langle S_R, M_R \rangle \) is the robot's description and \( D_H = \langle S_H, M_H, B_H \rangle \) is the human's description. Here, \( S \) defines types, predicates, and $pvariables$, while \( M \) specifies actions, preconditions, and effects. For instance, actions like {\small\texttt{human\_pick}} and {\small\texttt{robot\_pick}} modify the state fluent {\small\texttt{obj-loc}}. The model predicting human behavior \( B_H \) is derived from simulations with added noise impacting state transitions (see Section~\ref{sec:formulation-behavior} below). A task instance \( T = \langle O, I, G \rangle \) consists of objects \( O \), the initial state \( I \), and the goal state \( G \). We use the PROST planner~\cite{keller-eyerich-icaps2012} to compute an action sequence \( \pi = \langle a_1, \dots, a_K \rangle \) that transitions the system from \( I \) to \( G \) as a combination of actions to be executed by the robot and the human. This plan computation using a heuristic tree search method that maximizes expected cumulative rewards.



\subsection{Modeling Human Behavior as State Transitions}
\label{sec:formulation-behavior}
The model \(B_H\) of human behavior captures the uncertainty in the human's execution of specific actions. Since we were using a simulation environment for experimental evaluation, we had to simulate such uncertainty using empirical probability distributions not known to the robot. Specifically, we introduced noise by sampling from a Gaussian distribution centered on expected outcomes. 
The magnitude of this noise was adjusted automatically based on task complexity and considered different human behaviors, 
modeling execution failures as thresholds driven by specific criteria, e.g., a human's attempt to lift an object fails because they are not able to exert sufficient force. As a result, we were able to realistically simulate variability in human action execution, as discussed further in Section~\ref{41}.

\vspace{-0.5em}
\noindent
Once the noise distributions were determined, observations from 10 simulated trials of each of 11 cooking and cleaning tasks were used to learn an initial model of \(B_H\) as the state transition probabilities \(P_H(s' | s, a_H)\) for any particular human action \(a_H\). 
These probabilities were refined over subsequent trials, allowing the robot to predict human action outcomes more accurately. 
These probabilities were encoded in the domain file, and used to consider uncertainty in human action outcomes during planning and execution.

\subsection{Anticipation and Collaboration}
\label{sec:forumalation-anticipate}
\begin{figure}[h]
    \centering
\includegraphics[width=0.5\textwidth]{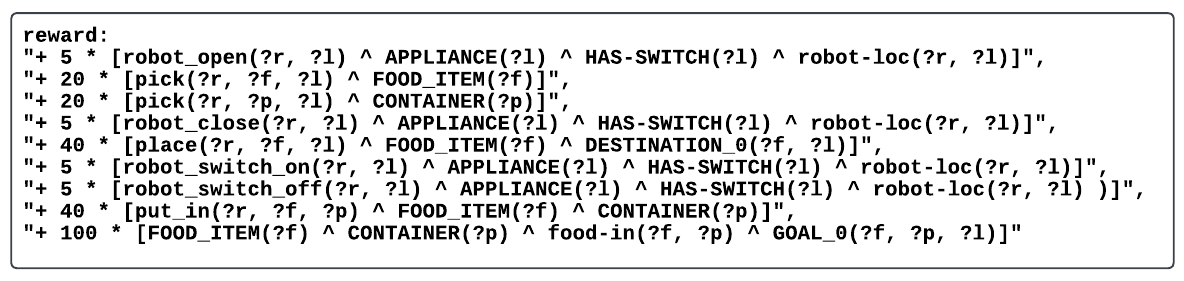}
    \caption{Partial description of reward specification for the $PrepareBreakfast(Toast)$ task.}
    \label{fig:rewards}
    \vspace{-1em}
\end{figure}

A key component of our framework is the reward specification that helps the robot trade off between completing the task successfully and the effort involved in preventing or recovering from potential failures in human action execution. 
Figure~\ref{fig:rewards} shows a simplified version of our reward function for a specific task (\textit{prepare breakfast}) based on subgoals that guide the robot toward task completion. Each component of the reward function models 
different interactions illustrated here in the context of tasks in the kitchen:
\begin{itemize}
\vspace{-0.3cm}
    \item \textbf{Appliance interaction:} successfully performing valid actions, e.g,, {\small\texttt{open}}, {\small\texttt{close}}, and {\small\texttt{robot\_switch\_on}} on appliances is rewarded. 

    \item \textbf{Item collection:} preparing for tasks, e.g., picking up {\small\texttt{FOOD\_ITEM}} or {\small\texttt{CONTAINER}} early, is rewarded.

    \item \textbf{Container placement:} placing objects at designated locations, e.g., containers such as plates and bowls in their {\small\texttt{DESTINATION}}, is rewarded.

    \item \textbf{Intermediate placement:} placing objects in intermediate locations in preparation for specific tasks, e.g., stove or toaster for cooking, is rewarded.



    \item \textbf{Final delivery:} placing an object in its correct goal location is rewarded.

    \item \textbf{Goal fulfillment:} satisfying all conditions of the {\small\texttt{GOAL}} receives a high reward.
    \vspace{-1em}
\end{itemize}
Similar reward functions are populated automatically for other actions and tasks in the domain.

\vspace{-0.5em}
\noindent
We also define a set of rewards that promote anticipatory and cooperative behaviors by aligning action choices with capabilities. We illustrate this in the context of humans performing pickup actions that are likely to result in failure, and the actions used to avoid or recover from these failures.
\begin{itemize}
\item \textbf{Reward for preventing failure.} This term rewards the robot for performing an action instead of a human in an attempt to prevent failure, e.g., pick up fragile items that a human may potentially drop and break.
\begin{align*}
R_{1} &= \sum_{\substack{r:\text{robot},\\ i:\text{item},\\l:\text{location}}}
\Big(\text{pick}(r,i,l) \wedge \text{fragile}(i) \nonumber\\
&\quad\wedge \,\exists h: \text{human-loc}(h,l)\Big)
\end{align*}

\item \textbf{Reward for preparing to respond to failure.} This term incentivizes the robot to place items that enable it to respond to failure near locations where a failure may occur, e.g., place a mop near location of fragile items that a human may pick up and potentially drop.
\begin{align*}
R_{2} &= \sum_{\substack{r:\text{robot},\\ m:\text{item},\\ l:\text{location}}}
\Big( \text{mop}(m) \wedge \text{obj-loc}(m,l) \wedge \text{robot-loc}(r,l) \nonumber\\
&\quad\wedge \exists h:\text{human-loc}(h,l) \nonumber\\
&\quad\wedge \exists x:\text{fragile}(x) \wedge \text{obj-loc}(x,l) \Big)
\end{align*}

\item \textbf{Penalty for not being prepared for failure.} This term penalizes the robot when a human performs an action that may lead to failure and the robot has not prepared for it accordingly, e.g., human allowed to pick up a fragile item without a mop nearby.
\begin{align*}
R_{3} &= -\sum_{\substack{l:\text{location},\\ i:\text{item}}}
\Big( \exists h:\text{fragile}(i) \wedge \text{pick\_human}(h,i,l) \nonumber\\
&\quad\wedge \,\nexists m: \text{mop}(m) \wedge \text{obj-loc}(m,l) \Big)
\end{align*}
\end{itemize}
Rewards can be suitably defined for other actions and domains using knowledge of actions likely to result in failures or help recover from failures. 
Each action executed by the robot also comes with a cost (i.e., negative reward) based on the effort needed (e.g., time spent, distance traveled) in completing the action. This allows the robot to trade off the need to prevent (or recover from) failures and the effort involved in achieving it. 

\section{Experimental Setup and Results}
\label{sec:exp}

We experimentally evaluated two hypotheses related to the performance of our framework.
\begin{enumerate}
\vspace{-0.3cm}
\item[\textbf{H1:}] Reasoning with learned or encoded models of human behavior improves performance and collaboration with a human compared with not using such models.
\item [\textbf{ H2:}] Incorporating the strategy to anticipate failures enables the robot to recover better from such failures compared with not using the strategy.
\vspace{-0.3cm}
\end{enumerate}
The experimental setup used for evaluation and the corresponding results are discussed below.

\subsection{Experimental Setup}
\label{41}
Our experimental setup involved three key components: learning a stochastic human behavior model, encoding domain knowledge in RDDL, and selecting appropriate baselines and evaluation measures.

\begin{figure}[tb]
\centering
\setlength{\belowcaptionskip}{-10pt}
\includegraphics[width=0.49\textwidth]{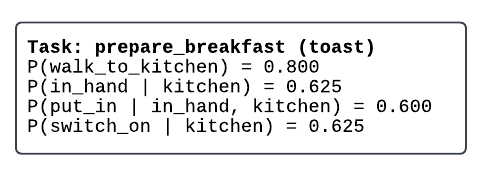}
\vspace{-2.5em}
\caption{Example probabilities of state transition probabilities in human behavior model. The probability of the human: walking to the kitchen from a random location is 0.8; grabbing a bread slice while in the kitchen is 0.625; placing the bread slice in the toaster after grabbing it is 0.6; switching on the toaster while in the kitchen is 0.625.}
\label{fig:conP}
\end{figure}

\vspace{-0.5em}
\noindent
\textbf{Learning the Human Behavior Model.} As stated in Section~\ref{sec:formulation-behavior}, we learned a probabilistic state transition model of human behavior in the VirtualHome simulator by decomposing tasks into actions and introducing noise sampled from a normal distribution ($\mu = 0, \sigma = 0.1$) filtered with a $0.5\sigma$ threshold. For each task, we ran 10 noisy simulations to compute conditional probabilities over state transitions. These probabilities also represent preferences (e.g., choosing fragile instead of non-fragile items) and deviations (e.g., leaving a room mid-task). Figure~\ref{fig:conP} shows probabilities of outcomes of the human's actions related to a particular task. 

\vspace{-0.5em}
\noindent
\textbf{Encoding Planning Models.}  
We modeled the environment using RDDL as described in Section~\ref{sec:formulation-taskplan}. The domain included 11 generic household (cooking and cleaning) tasks based on nine food items, eight appliances, nine cutlery items, and five cleaning items. Human actions were treated as exogenous stochastic transitions that (unknown to the robot) were based on the learned model. 
We used a predefined JSON file to map natural language LLM outputs into RDDL-specific syntax for goals and rewards.

\vspace{-0.5em}
\noindent
We used the PROST Planner~\cite{keller-eyerich-icaps2012}, with the Trial-based Heuristic Tree Search algorithm on a Factored MDP, integrating Upper Confidence Bound for action selection, Unsolved Monte Carlo for handling uncertainty, Partial Bellman Backup for Q-value estimation, and Iterative Deepening Search for heuristic Q-value initialization. We balanced exploration and exploitation by combining the IPC2011 configuration that supports broad exploration and the IPC2014 configuration that improves decision-making by prioritizing informative samples. We set a maximum planning horizon of 60 for the joint goal (two tasks) to limit search. Our reward function rewards task completion and progress, penalizes delays, and discourages failures (Section~\ref{sec:forumalation-anticipate}). The planner generated joint-action (human and robot actions) sequences that optimize:
(a) \textit{distance to target}: nearest agent handles the object; (b) \textit{action prioritization}: robot chooses to perform interactions with fragile objects; (c) \textit{goal relevance}: relevant \textit{object types} are considered; and (d) \textit{plan length}: minimal plan is computed. Recall that human actions are modeled and observed but are not determined by our framework.


\vspace{-0.5em}
\noindent
\textbf{Performance Measures and Baselines.}  
To evaluate \textbf{H1} and \textbf{H2}, we performed 30 simulation rollouts, each with five collaborative tasks in our household domain. Each task was a high-level goal, such as "Prepare and Serve Toast and Coffee." This task can be split into subgoals, like "Toast the Bread," and "Place items on the table." To achieve each (sub)goal, the robot had to execute a sequence of actions such as "Open Toaster," "Put bread in Toaster, and " "Switch on Toaster." Each simulation involved multiple potential failures in the form of undesired human action outcomes. We used the following performance measures:
\begin{itemize}
\vspace{-0.3cm}
    \item \textbf{Average number of actions}: the mean number of actions for completing entire task(s). 
    
    \item \textbf{Number of failures}: count of failures during execution of the assigned task(s).
    
    \item \textbf{Number of failures prevented}: count of instances of robot preventing failure in human action outcomes.
    
    \item \textbf{Number of failures recovered}: count of instances in which robot recovered from failure by anticipating it. 
    
    \item \textbf{Task completion rate}: fraction of tasks completed successfully; higher value indicates better performance.
    
    \item \textbf{Subgoal completion rate}: fraction of subgoals achieved, measuring adherence to task(s).
\vspace{-0.5em}
\end{itemize}
Each simulation included robot and human actions such as {\small\texttt{pick}}, {\small\texttt{place}}, {\small\texttt{move}}, {\small\texttt{switch}}, {\small\texttt{open}}, and {\small\texttt{put\_in}}. The robot evaluated multiple trajectories, optimizing for expected reward. The environment includes fragile and non-fragile objects (e.g., fruits, cereal, mop, bread, milk) and dynamic constraints in the form of obstructions and unavailable items. As summarized later (in Table~\ref{tab:prepare_serve_recovery_all}), we assessed the robot's ability to: (a) anticipate human delays or mistakes; (b) recover from missteps such as tool misuse or incorrect object selection; and (c) adapt to constraints based on rewards that consider plan length and failed subgoals.

\begin{table*}[tb]
    \centering
    \renewcommand{\arraystretch}{1.2}
    \setlength{\tabcolsep}{6pt}
    \begin{tabular}{|lccc|ccc|}
        \toprule
        \textbf{Task} &
        \multicolumn{3}{c|}{\textbf{Subgoals completion \%}} & 
        \multicolumn{3}{c|}{\textbf{Task completion \%}} \\
        & \textbf{(LLM)} & \textbf{(RDDL)} & \textbf{(Ours)} & \textbf{(LLM)} & \textbf{(RDDL)} & \textbf{(Ours)} \\
        \midrule
        Prepare and Serve Salmon + Water     & 46.67\% & 63.3\%  & \textbf{85\%}    & 30\% & 55\%   & \textbf{80\%} \\
        Prepare and Serve Coffee + Wash Dish & 47.5\%  & 75\%    & \textbf{86.7\%}  & 20\% & 55\%   & \textbf{87.5\%} \\
        Prepare and Serve Cereal + Coffee    & 52.5\%  & 70\%    & \textbf{88.3\%}  & 25\% & 60\%   & \textbf{85\%} \\
        Prepare and Serve Toast + Coffee     & 42.5\%  & 68.3\%  & \textbf{83.3\%}  & 15\% & 58.3\% & \textbf{84\%} \\
        Prepare and Serve Pizza + Wash Dish  & 35.56\% & 66.7\%  & \textbf{84.2\%}  & 10\% & 57\%   & \textbf{83.4\%} \\
        \midrule
        \textbf{Average\%} & 44.55\% & 68.26\% & \textbf{85.5\%} & 20\% & 57.06\% & \textbf{84.78\%} \\
        \bottomrule
    \end{tabular}
    \caption{Task and subgoal completion performance of our framework is substantially better than that of the two baselines, LLM-only and RDDL/PROST without anticipatory rewards, over selected composite tasks.}
    \label{tab:prepare_serve_eval}
\end{table*}

\begin{table*}[tb]
    \centering
    \renewcommand{\arraystretch}{1.2}
    \setlength{\tabcolsep}{4pt}
    \begin{tabular}{|l|ccc|ccc|ccc|ccc|}
        \toprule
        \textbf{Task} & 
        \multicolumn{3}{c|}{\textbf{Failures}} & 
        \multicolumn{3}{c|}{\textbf{Prevention}} & 
        \multicolumn{3}{c|}{\textbf{Recovery}} & 
        \multicolumn{3}{c|}{\textbf{Avg. Actions}} \\
        & \textbf{L} & \textbf{R} & \textbf{O} & 
          \textbf{L} & \textbf{R} & \textbf{O} & 
          \textbf{L} & \textbf{R} & \textbf{O} & 
          \textbf{L} & \textbf{R} & \textbf{O}\\ 
        \midrule
        Prepare and Serve Salmon + Water     & 18/30 & 17/30 & 14/30 & 12/30 & 13/30 & 16/30 & 6/18 & 0 & 11/14 & -- & -- & 38 \\
        Prepare and Serve Coffee + Wash Dish & 9/30  & 11/30 & 8/30  & 21/30 & 19/30 & 22/30 & 0  & 2/11 & 6/8   & 24  & -- & 26 \\
        Prepare and Serve Cereal + Coffee    & 25/30 & 13/30 & 10/30 & 5/30 & 17/30 & 20/30 & 9/25 & 0 & 8/10  & --  & -- & 42  \\ 
        Prepare and Serve Toast + Coffee     & 27/30 & 9/30 & 12/30 & 3/30 & 21/30 & 18/30 & 6/27 & 0 & 9/12  & --  & -- & 48  \\
        Prepare and Serve Pizza + Wash Dish  & 18/30 & 11/30 & 9/30  & 12/30 & 19/30 & 21/30 & 3/18 & 0 & 7/9   & --  & -- & 30  \\ 
        \midrule
        \textbf{Average} & 20.0 & 12.2 & 10.6 & 10.6 & 17.8 & 19.4 & 4.8 & -- & 8.2 & -- & -- & 36.8 \\ 
        \bottomrule
    \end{tabular}
\caption{Failure prevention and recovery statistics for selected composite tasks. \textit{Prevention} indicates proactive avoidance of likely human failures, while \textit{Recovery} refers to corrective action after failure has occurred. In most scenarios, the LLM-only baseline (\textbf{L}) or RDDL/PROST baseline (\textbf{R}) failed to complete the composite tasks, with \textit{Avg. Actions} and \textit{Time Taken} reported as `--'. Our framework (\textbf{O}) consistently completes tasks through anticipation and recovery.}
\label{tab:prepare_serve_recovery_all}
\vspace{-2em}
\end{table*}

\vspace{-0.5em}
\noindent
As \textbf{baselines} for comparison, we considered just the LLM (\textbf{L}) and just the RDDL-based planner (\textbf{R}). Similar to existing literature~\cite{10802284}, the LLM baseline directly computed a sequence of actions for the joint goal (current and predicted next task). 
The RDDL-based baseline did not include the predictive model of human behavior in its reward specification; its reward structure directed the robot to follow a goal-conditioned plan to complete the task(s). 

\subsection{Experimental Results}
\label{subsec:exp_results}

\vspace{-0.25em}
\noindent
\textbf{Table~\ref{tab:prepare_serve_eval}} and \textbf{Table~\ref{tab:prepare_serve_recovery_all}} summarize results for a set of representative composite tasks. Our framework (\textbf{Ours, O}) consistently outperformed the two baselines: RDDL/PROST without anticipatory rewards (\textbf{RDDL, R}) and LLM-only plans (\textbf{LLM, L}). For example, in \textbf{Table~\ref{tab:prepare_serve_eval}}, average subgoal completion rate across all tasks was 85.5\% for our framework, compared with 68.26\% for the RDDL baseline and 44.55\% for the LLM baseline; substantial performance improvements were observed for overall task completion as well. Also, there was considerable variation in the results obtained with the baselines depending on the type and complexity of the tasks, whereas the performance remained consistently good with our framework. In addition, the task (or subgoal) completion rate was not $100\%$ with our framework only because the robot ran out of time to complete the tasks; this limit was relaxed for the numbers shown in \textbf{Table~\ref{tab:prepare_serve_recovery_all}} below. These results indicate the importance of the human behavior prediction model and the reward mechanism that leverages this model. 

\begin{figure}[tb]
\centering
\setlength{\belowcaptionskip}{-10pt}
\includegraphics[width=0.49\textwidth]{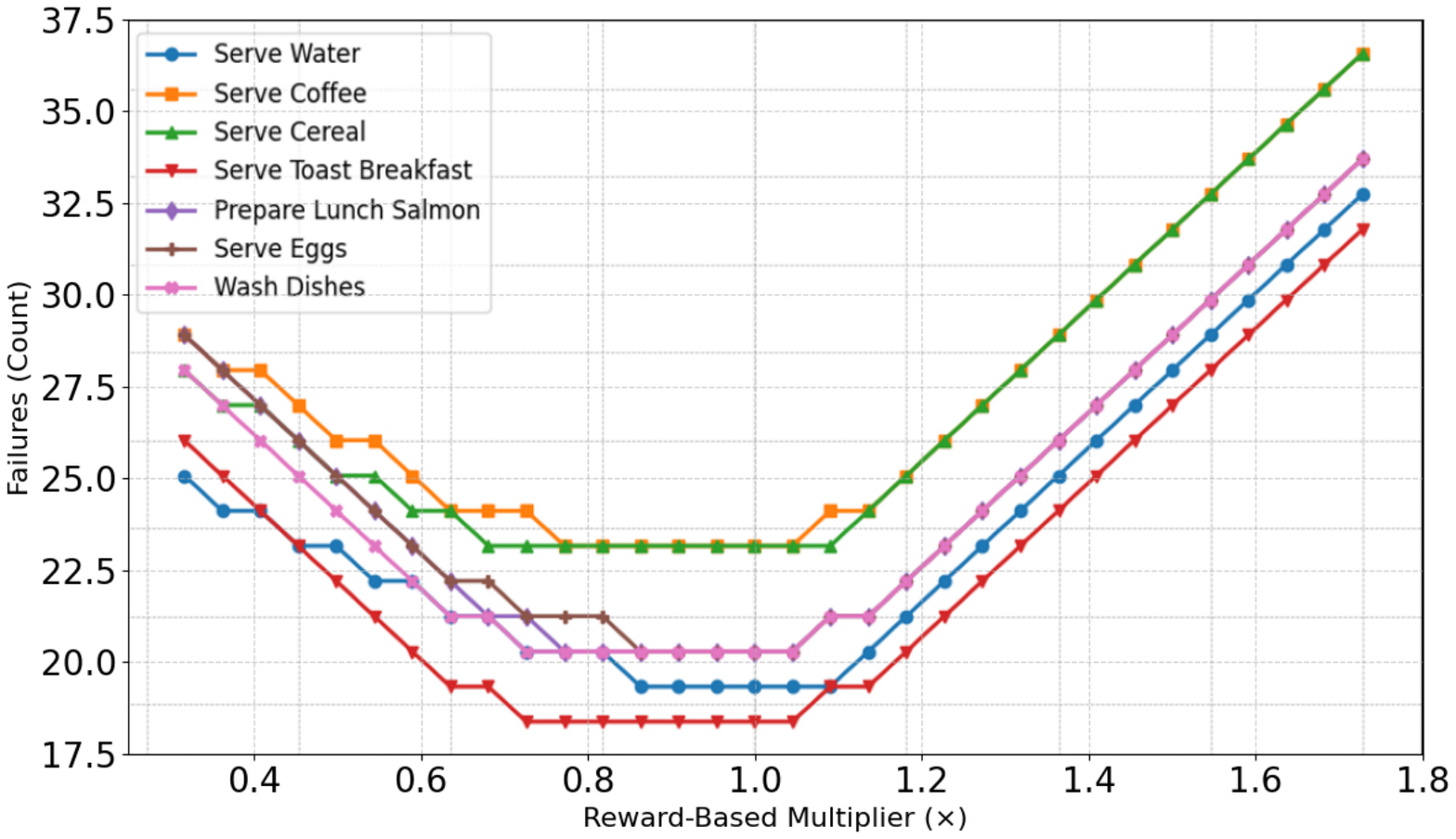}
\vspace{-2.5em}
    \caption{Task failures as a function of the reward-based multiplier that determines extent to which failure prevention and recovery is prioritized over task completion. 
    With an increase in the multiplier, failures reduce up to a point before increasing again. A trade off between failure prevention and task completion leads to good performance.}
    \label{fig:rewards_failures_graph}
\vspace{-1em}
\end{figure}

\vspace{-0.5em}
\noindent
\textbf{Table~\ref{tab:prepare_serve_recovery_all}} summarizes statistics of failures, failure prevention, and recovery across tasks. The robot prevented failures in many more trials when using our framework than when it used the baselines; it also recovered from a larger percentage of the errors that were not prevented. 
Recall that neither baseline explicitly reasons about potential failures to prevent or recover from them. Any prevention of failure or recovery from failure with our baselines was hence purely happenstance, e.g., the robot using the RDDL baseline chose to fetch the glass of water that a human may have dropped, or the LLM baseline happened to direct the robot to clear the mess created by the glass of water dropped by the human. As observed in \textbf{Table~\ref{tab:prepare_serve_eval}}, there was variation in the performance of the baselines based on the type and complexity of the tasks, but the performance remained consistent with our framework. Furthermore, our framework resulted in tasks being completed in each trial, including when the robot did not prevent or recover from errors. With the two baselines, on the other hand, the tasks remained incomplete in most trials. 
These results indicate the robot's ability to collaborate effectively with the human; it anticipated and prevented failures, recovered from failures, and/or found other plans to complete the task, thus supporting \textbf{H1} and \textbf{H2}.

\vspace{-0.5em}
\noindent
We also explored the trade off between  task completion and failure prevention or recovery. Figure~\ref{fig:rewards_failures_graph} presents the number of failures for different tasks and different reward multipliers; for a higher value of the multiplier, the reward structure assigns higher weight to failure prevention and recovery. The sensitivity of performance to this multiplier depended on the type and complexity of the task, 
but the increased focus on failures improved performance up to a point before having a negative impact. These results suggest that our reward structure can be adapted to the tasks and domain.


\section{Conclusion}
\vspace{-0.1cm}
This paper described a hybrid framework that enables a robot collaborating with a human to anticipate upcoming tasks, reason about potential failures due to actions executed by the human based on a learned model of their action capabilities, and to act to prevent these failures or to recover from them.
The framework combines the complementary strengths of knowledge-based and data-driven methods. In particular, it combines the LLM-based statistical prediction and the RDDL-based probabilistic relational sequential decision making capabilities. Experimental results demonstrate the substantial improvement in performance compared with LLM-based and RDDL-based baselines. Future work will explore the use of this framework on a physical robot and its extension to support multiagent collaboration.  





\balance

\bibliographystyle{IEEEtran}
\bibliography{IEEEabrv,references}
\end{document}